# Methodology to Deploy CNN-Based Computer Vision Models on Immersive Wearable Devices

Kaveh Malek[1], Fernando Moreu[2], [1] Department of Mechanical Engineering, [2] Department of Civil, Construction and Environmental Engineering, University of New Mexico, New Mexico

ABSTRACT

Convolutional Neural Network (CNN) models often lack the ability to incorporate human input, which can be addressed by Augmented Reality (AR) headsets. However, current AR headsets face limitations in processing power, which has prevented researchers from performing real-time, complex image recognition tasks using CNNs in AR headsets. This paper presents a method to deploy CNN models on AR headsets by training them on computers and transferring the optimized weight matrices to the headset. The approach transforms the image data and CNN layers into a one-dimensional format suitable for the AR platform. We demonstrate this method by training the LeNet-5 CNN model on the MNIST dataset using PyTorch and deploying it on a HoloLens AR headset. The results show that the model maintains an accuracy of approximately 98%, similar to its performance on a computer. This integration of CNN and AR enables real-time image processing on AR headsets, allowing for the incorporation of human input into AI models.

## I. INTRODUCTION

Wearable immersive technology has made significant progress in recent years, particularly in its applications for real-time visual tasks. Yet, the current AR headsets are limited by their processing power, preventing researchers from integrating their visual capabilities with computationally heavy algorithms such as image recognition based on CNN models [1]. CNN is widely accepted for its superior performance in image recognition and classification tasks [2]. Despite this superiority, their integration into AR headsets remains challenging due to the computational demands of these models during the training and evaluation process. Additionally, the C#-UnityEngine platform used for programming AR headsets does not support image-processing libraries such as OpenCVSharp and EmguCV [3], nor does it support matrix computation libraries like MathNet for deployment on AR headsets [4]. The integration of CNN models with AR wearable platforms requires addressing the gap between the complexity of CNN models and the processing limitations of AR headsets, as well as the simplicity of the available functions in programming platforms for AR wearable devices.

Researchers and industry have developed CNN models for hand-held AR devices such as cell-phones. For instance, researcher at the University of Hong Kong designed a neural network suitable for mobile and embedded devices with limited computing power called ShuffleNet [5]. DeepScale at the University of California, Berkeley, and Stanford University developed SqueezeNet [6] for mobile systems, which included a compound scaling technique to balance efficiency and accuracy in deep learning. Google Brain created specific architectures for mobile applications such as MobileNet [7], NASNet [8], and EfficientNet [9]. Others have finetuned existing CNN models, such as MobileNet [10] and SqueezeNet [11], or combined models like MobileNet with Yolo-family [12] to create AR applications on smartphones. These efforts demonstrate the potential of CNNs in portable devices, but they do not directly address the specific limitations of AR headsets.

The accomplished approach to integrating CNN models on AR headsets typically involves running the models on stationary devices, such as servers or desktop computers, and then transmitting the results to AR wearables via wired or Wi-Fi connections. This approach falls under the Internet-of-Things (IoT) paradigm, where interconnected devices communicate and share data. The IoT approach uses the processing power of stationary devices to perform complex CNN calculations and then sends the results to the AR headset for display and interaction [13], [14], [15], [16], [17]. This method relies on a consistent network connection and does not fully utilize the standalone capabilities of AR headsets.

Some research has successfully implemented rule-based or pattern recognition AI models directly within AR headsets, proposing standalone approaches that eliminate the need for external processing. Some rule-based algorithms that are implemented in AR headsets include Canny edge detection [4], a crack-measurement algorithm [3], and an automatic Region Of Interest (ROI) selection [18]. Unlike the IoT approach, which relies on stationary devices for processing, these standalone methods enable the AR headsets to perform computations independently. However, these models are typically less complex than CNNs and do not offer the same level of accuracy or flexibility.

This study presents a method for deploying CNN models on AR headsets by first training the models on more powerful computers and then transferring the optimized weight matrices to the headset. Our approach involves transforming the image data and CNN layers into a one-dimensional format suitable for the AR platform, effectively addressing the processing power limitations. We demonstrate this methodology using the LeNet-5 CNN model, trained on the MNIST dataset with PyTorch and deployed it on a HoloLens AR headset. The results show that the model achieves an accuracy of approximately 98%, comparable to its performance on a computer. This integration of CNN and AR achieves real-time image processing on AR headsets and thereby enables human input into AI models.

## II. METHODOLOGY

This section outlines the methodology for integrating CNNs within the Unity3D engine to ensure deployment on AR headsets. We start with an overview of the AR headset platform, describing Unity3D as the primary environment for developing AR applications. Then we outline the methodology which includes steps for offline training and online implementation. Finally, it describes the multi-stage approach for transforming and executing CNN models in real-time on AR devices, ensuring efficient performance despite

### A. AR Headset Platform Overview

This study proposes a methodology for the integration of CNNs in the Unity3D engine so that the result is deployable in AR headsets. Unity3D is the primary platform for creating the AR application because it includes a comprehensive simulation environment for AR development where developers can flexibly prototype and interact with the virtual world through user-defined scripts. This environment is supported by the main Unity3D library, UnityEngine, which is essential for all Unity scripts. The core scripting language used in Unity is C#, enabling the developers to define the behavior of game objects and manage the AR environment. Additionally, the Universal Windows Platform (UWP) is employed to build and deploy applications on Windows-based AR devices, ensuring compatibility and streamlined deployment processes. Finally, developers can use AR SDK which provides the tools and APIs for AR development.

Despite these advantages, the AR platform presents significant challenges, particularly concerning library inconsistencies. Libraries for advanced mathematics, matrix computation, and image processing—such as EmguCV, Magick.NET, OpenCVSharp, and ImageProcessor—are not fully compatible with the C# interface for AR headset programming. This inconsistency necessitates additional preprocessing steps and transformations to ensure compatibility. Furthermore, processing 2D images within a platform primarily designed for 1D mathematical operations introduces complexity. The limited processing capabilities of AR headsets also constrain the real-time execution of computationally intensive CNN models. A viable approach to implementing CNNs in AR headsets would involve addressing these challenges and limitations.

### B. Outline of Model Training and Implementation

Figure 1 outlines the training and implementation process of deploying CNNs in AR headsets. The model training and implementation process is divided into offline and online steps. Initially, the CNN model is trained on a stationary processing unit such as a computer to use its superior processing capabilities. After training, we extract the optimized weight matrices from each convolutional layer, representing the learned features used for inference in the online stage. The connections between neurons in different layers are identified, to make sure the model architecture remains intact during the transformation process. These weights and connections are then transformed into a format suitable for deployment on the AR platform. This involves converting them into a one-dimensional format compatible with the AR environment. Next, the transformed weights and connections are deployed on the AR headset.

Once deployed, the online steps ensure the model runs in real-time on the AR headset. In the first online step, the AR device's camera captures input images at user request, which are then transformed into a one-dimensional array and fed into the CNN model for processing. After processing images with the CNN model, the output of the model predictions is presented to the user through the AR headset, either visually or audibly.

These steps collectively address the computational limitations of AR headsets while enabling the integration of advanced CNN models for real-time image recognition tasks. In fact, the processing power of the headset, although limited, is sufficient for executing the transformed CNN model due to the preprocessing done during the offline steps.

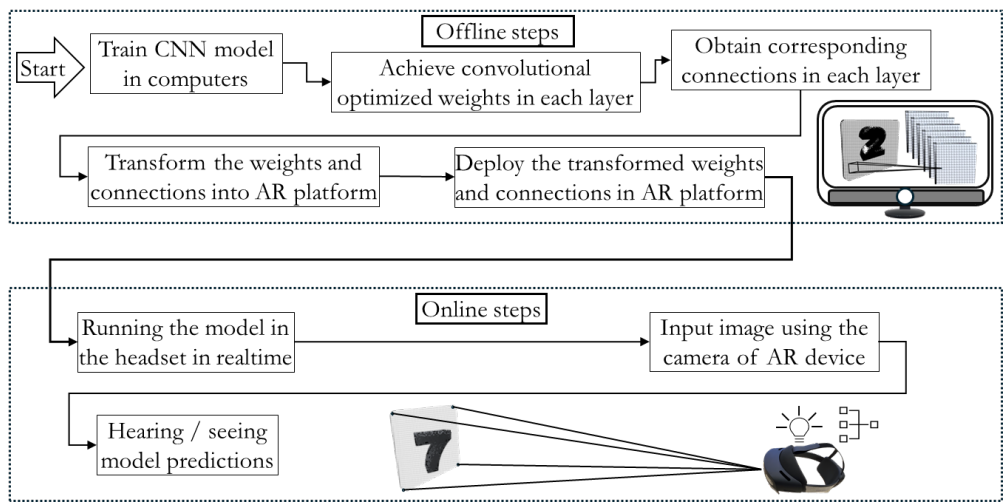

Figure 1. Online and offline steps, which are used for training and implementation.

*C. Integration of CNNs in AR Platform*

Deploying CNNs on AR headsets requires a multi-stage approach. Figure 2 illustrates this methodology, by dividing it into four main panels, each detailing a specific aspect of the integration process: 1. offline training on computers, 2. acquiring optimized parameters, 3. transforming these parameters for the AR platform, and 4. executing the model in real-time on AR images.

The initial step is the offline training of the CNN model on a stationary computer as shown in panel 1. This process begins with running the CNN model, where input images are processed through various convolutional layers to generate predictions. These predictions are then compared to the ground truth labels, and the loss or error is calculated based on the discrepancy between the predicted and actual labels. Backward propagation is subsequently performed to optimize the parameters of the model, adjusting the weights and biases to minimize the error. This cycle of forward pass, loss calculation, and backward propagation is repeated over multiple epochs, with the model evaluated periodically to monitor improvements in accuracy and performance.

Once the training phase is complete, the optimized parameters, including the weight matrices and biases, are extracted. This involves identifying the connections between neurons in successive layers and the specific weights applied to these connections. For instance, the elements of the optimized weight matrix and bias from the $l^{th}$ channel of the $k^{th}$ layer, which forms the $r^{th}$ channel of the $(k+1)^{th}$ layer, are shown in panel 2 of Figure 2. The next step involves transforming these optimized parameters into a format suitable for the AR platform as described in panel 3. Given the limitations of AR headsets regarding one-dimensional (1D) processing capabilities, the 2D weight matrices and biases are converted into a 1D format. This transformation ensures compatibility with the AR environment while preserving the functional relationships between layers. The detailed process includes mapping the 2D convolutional operations into equivalent 1D operations that the AR platform can execute efficiently. This step is critical to overcoming the inherent limitations of the AR hardware and making advanced CNN functionalities feasible.

The final step, as illustrated in panel 4, is the real-time execution of the transformed CNN model on the AR headset. The AR device's camera captures input images, which are then processed by the transformed model. The images are first converted into a 1D array compatible with the AR platform's processing capabilities. The CNN model processes these arrays, utilizing the transformed weights and biases to generate the transformed data for each channel and eventually make predictions. For example, the information in the $l^{th}$ channel in the $k^{th}$ layer processes the input using the previously transformed weight matrix and bias, producing the data of $r^{th}$ channel in the $(k+1)^{th}$ layer as shown in panel 4 of Figure 2. In one-dimensional arrays, data processing for CNNs begins with convolving image pixels with corresponding weights and progresses through layers until reaching the final layer, which is often a fully-connected layer. The outputs are then rendered on the AR display, providing real-time visual or auditory feedback to the user.

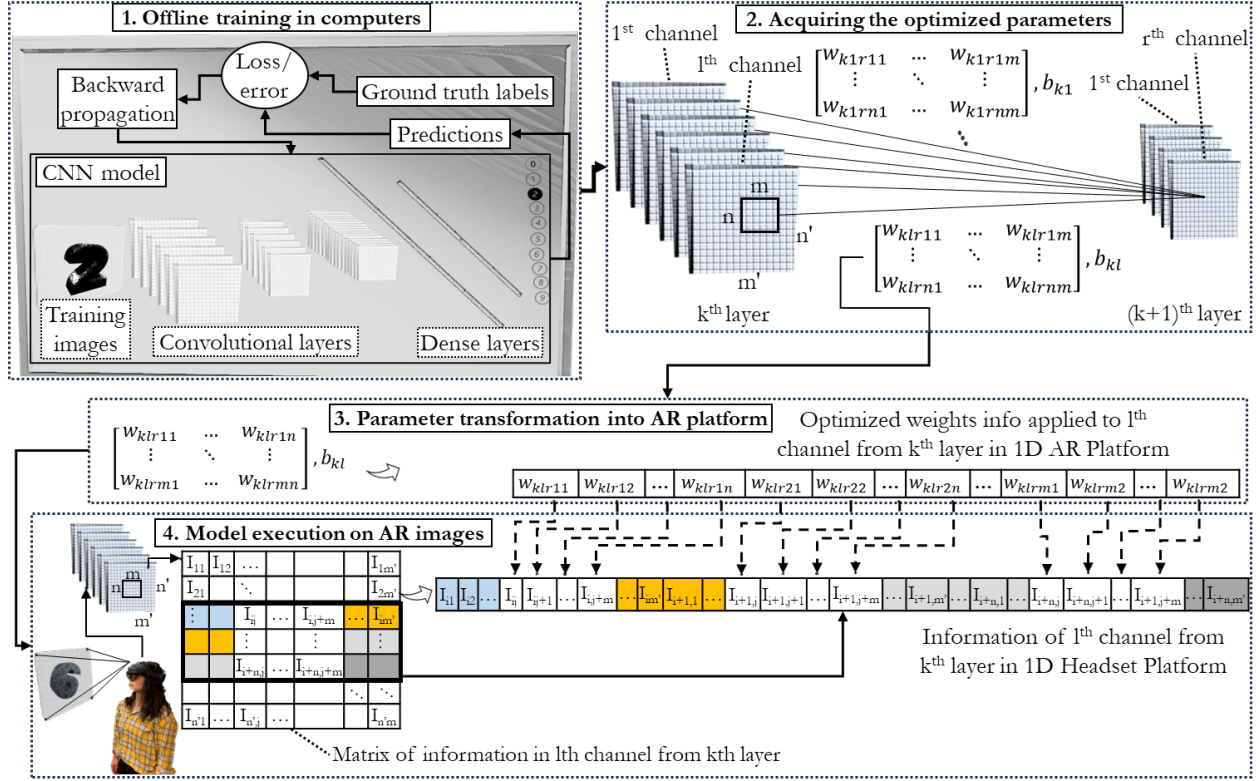

Figure 2. Method for implementing CNN-based computer vision models on immersive wearable devices.

## III. IMPLEMENTATION

This section details the integration of the LeNet-5 CNN into the AR platform for real-time digit recognition. It begins with an overview of the LeNet-5 architecture, highlighting its historical significance and structural components. This is followed by the process of training LeNet-5 using the MNIST dataset. The section then describes the steps taken to reshape weights and code the model in C#-UnityEngine for compatibility with AR headsets. Finally, it discusses testing the LeNet-5 model on MNIST digits using HoloLens 2.

### A. LeNet-5 Overview

LeNet-5 [19], developed by Yann LeCun and his collaborators in the late 1998s, is one of the earliest CNNs and has significantly contributed to the advancement of deep learning. It was originally designed for handwritten digit recognition on the MNIST dataset, which has become a benchmark in machine learning. The architecture of LeNet-5 is relatively simple yet powerful, consisting of seven layers including convolutional layers, subsampling (pooling) layers, and fully connected layers. This simplicity and effectiveness make it an ideal candidate for integration into the AR platform for real-time digit recognition and other applications.

The LeNet-5 architecture, as depicted in Figure 3, comprises two convolutional layers (C1 and C3) interspersed with two subsampling layers (S2 and S4), followed by two fully connected layers (C5 and F6), and finally an output layer (F7). The convolutional layers use learnable filters to extract features from the input images, while the subsampling layers reduce the spatial dimensions, thereby controlling overfitting and improving computational efficiency. Each convolutional layer is followed by an activation function (ReLU), which introduces non-linearity into the model. The fully connected layers further process the features extracted by the convolutional layers, ultimately producing the final classification result.

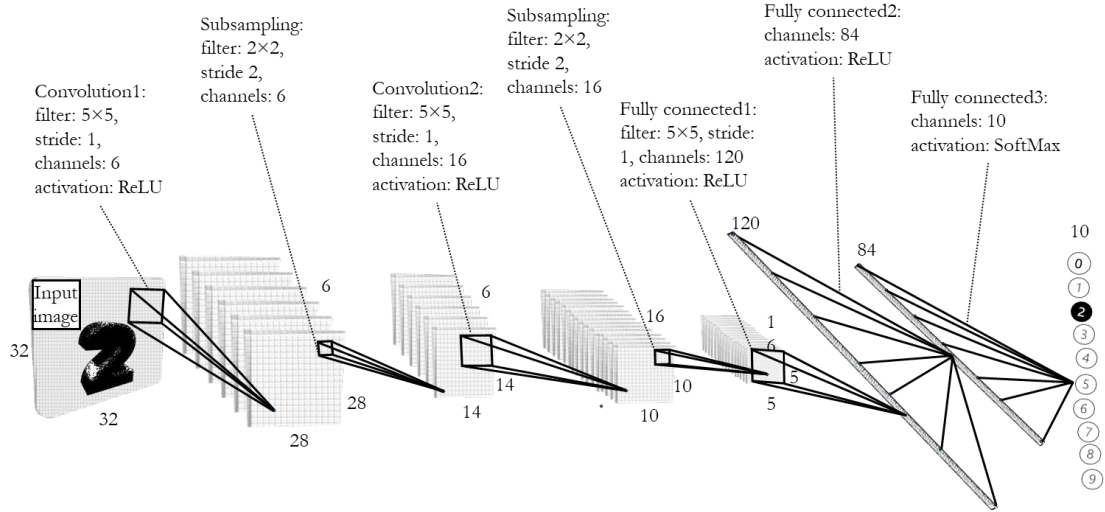

Figure 3. LeNet-5 architecture which is used to predict the MNIST in AR headsets.

*B. Integration Of LeNet-5 in AR Platform*

This part explains how LeNet-5 is adapted for use in AR headsets. It starts with training the model on the MNIST dataset and then describes the process of reshaping weights and coding in C#-UnityEngine for compatibility with AR headsets.

*a. Training with MNIST dataset*

The MNIST dataset, a large collection of handwritten digits, is a standard benchmark for evaluating image classification algorithms. It comprises 60,000 training images and 10,000 testing images, each of 28×28 pixels in grayscale. For training LeNet-5 on the MNIST dataset, we used a batch size of 32, a learning rate of 0.001, and trained the model over 10 epochs in PyTorch. The input image to the LeNet-5 model is 32×32 for digit recognition, with two rows and columns of padding added on each side for the training process. The training process involves feeding the input images through the convolutional layers, where filters learn to identify features such as edges and textures. After each epoch, the model's weights are updated to minimize the loss function, measured by the difference between the predicted and actual labels. As expected, by the end of the training phase, the model achieved high accuracy.

*b. Weight Reshaping and coding in C#-UnityEngine*

To integrate LeNet-5 into an AR platform, the optimal hyperparameters were saved in C# arrays. The weights and biases were reshaped to match the architecture requirements of the AR platform. The process involved converting the 2D convolutional operations into 1D arrays for compatibility with the AR headset's computational framework. Because the C#-UnityEngine platform used for programming AR headsets does not support advanced mathematics, matrix computation, and image processing, the coding process in the AR headset platform differs from traditional implementations on computers. As shown in Figure 4, we compared functions across PyTorch and the C#-UnityEngine environment. The first row of the figure compares normalization or scaling functions in PyTorch and C#-UnityEngine environment, while the second row shows the difference in the inner product of arrays between these two programming platforms. The third row shows the implementation of max-pooling with a kernel size of 2×2 and stride of 2, and the fourth row demonstrates the ReLU activation function.

| Function | PyTorch | C#-UnityEngine Environment |
|---|---|---|
| Scaling | torch.nn.functional.normalize | ```csharp
private static double[] NormalizeData(IEnumerable<double> data, double min, double max)
{
    double dataMax = data.Max();
    double dataMin = data.Min();
    double range = dataMax - dataMin;

    return data
        .Select(d => (d - dataMin) / range)   // Normalize to [0, 1]
        .Select(n => (1 - n) * min + n * max) // Scale to [min, max]
        .ToArray();
}
``` |
| Arrays inner product | numpy.dot | ```csharp
static double SumProduct(double[] array, double[] weights,
    int arrayStartIndex, int weightsStartIndex, int count)
{
    double sum = 0;
    for (int k = 0; k < count; k++)
    {
        sum += array[arrayStartIndex + k] * weights[weightsStartIndex + k];
    }
    return sum;
}
``` |
| Max-Pooling kernel:2×2 stride: 2 (the example is related to channel 0, layer 4) | torch.nn.functional.max_pool2d (x, 2, 2) | ```csharp
for (int i = 0; i < 28; i += 2)
{
    for (int j = 0; j < 28; j += 2)
    {
        l3c0[ind_i * 14 + ind_j] = Math.Max(l2c0[i * 28 + j], Math.Max(l2c0[i * 28 + j + 1],
            Math.Max(l2c0[(i + 1) * 28 + j], l2c0[(i + 1) * 28 + j + 1])));
``` |
| ReLU Activation (the example is related to the first convolution layer) | torch.nn.functional.relu | ```csharp
l2c0[index] = Math.Max(
    SumProduct(grey_image_normalized, conv1_weight_0, index_img, 0, 5) +
    SumProduct(grey_image_normalized, conv1_weight_0, index_img + 32, 5, 5) +
    SumProduct(grey_image_normalized, conv1_weight_0, index_img + 64, 10, 5) +
    SumProduct(grey_image_normalized, conv1_weight_0, index_img + 96, 15, 5) +
    SumProduct(grey_image_normalized, conv1_weight_0, index_img + 128, 20, 5) + conv1_bias[0], 0);
``` |

Figure 4. Comparison of the functions for implementing the CNN model in C#-UnityEngine environment with Python. Find the entire code inside the Scripts folder of the LeNetAR Unity project at https://github.com/KaMa85/LeNet.

## C. Testing LenNet on MNIST in AR Headsets

To evaluate the performance of LeNet-5 in an AR environment, we conducted an experiment using an AR headset developed by the Microsoft Corporation. The test setup involved a test operator wearing HoloLens 2 to predict MNIST digits printed and attached to a wall as shown in Figure 5. The operator viewed these images through the headset, which superimposed the predicted labels in front of him after receiving the user's voice command "predict". The operator was instructed to observe each MNIST digit from a consistent distance of approximately 1.5 meters, ensuring that the HoloLens 2 could capture clear and stable images for prediction. The digits were printed on paper and attached to a plain white wall. The experiment included a variety of digits from the MNIST dataset.

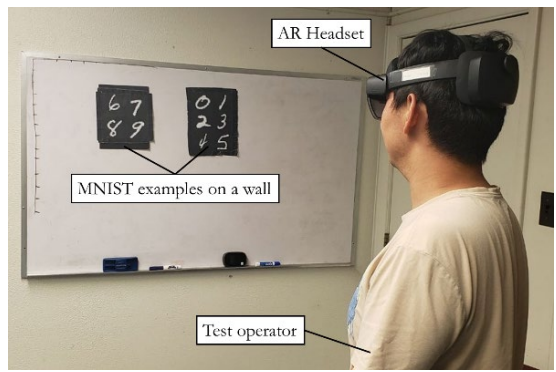

Figure 5. An instance of the experiment

In Figure 6, we can see the successful predictions made during the experiment. Each subfigure (a)-(h) illustrates MNIST digits from 1 to 8, attached to the wall, with the predicted labels superimposed by the HoloLens 2. The superimposed labels were generated by the AR platform's processing of the images captured by the HoloLens 2's front-facing camera, which then used the pre-trained LeNet-5 model to predict the digit in real-time.

We also observed instances of incorrect predictions caused by unstable image acquisition, which underscored areas for improvement in the AR system's image stabilization and recognition processes. Figure 7 presents two examples of such incorrect predictions. In the first case, shown in Figure 7(a), the top part of the digit resembled '1', while the bottom part resembled '3', leading to an incorrect prediction due to the partial overlap of two digits. This error occurred because the headset captured an image where two digits were incompletely framed, resulting in a composite image that confused the model. In the second example, Figure 7(b) depicts a digit '1' misrepresented as '7' due to a white horizontal line on the wall, which made '1' appear similar to '7'. This line introduced a visual artifact that the model misinterpreted as a significant feature, demonstrating the sensitivity of the recognition process to environmental factors and the need for better image preprocessing techniques to filter out such noise. The experiment demonstrated both the potential and challenges of implementing CNNs for object recognition in AR headsets.

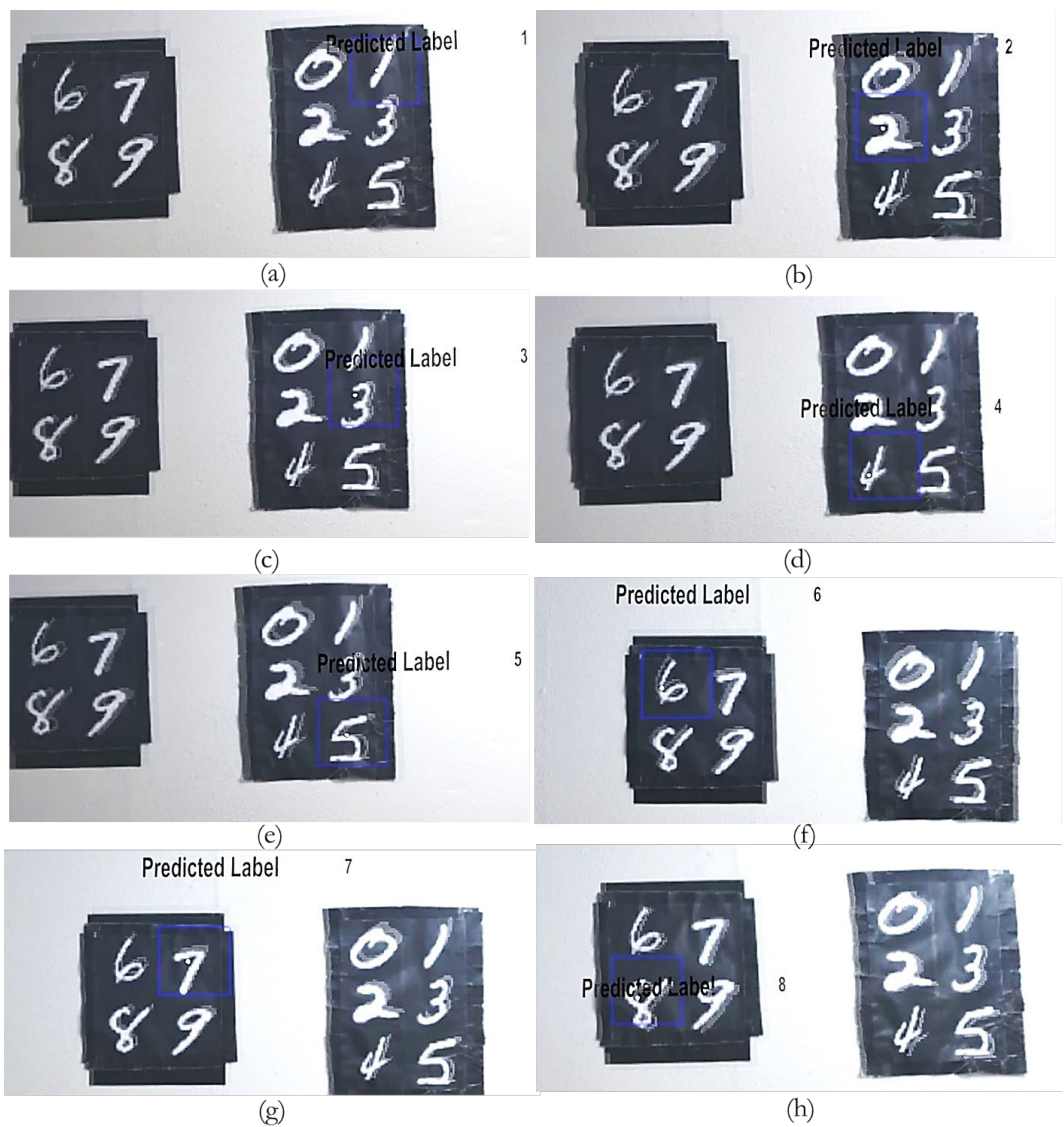

Figure 6. Example results of experiments (a)-(h) MNIST images from 1 to 8 on the wall with superimposed predicted labels.

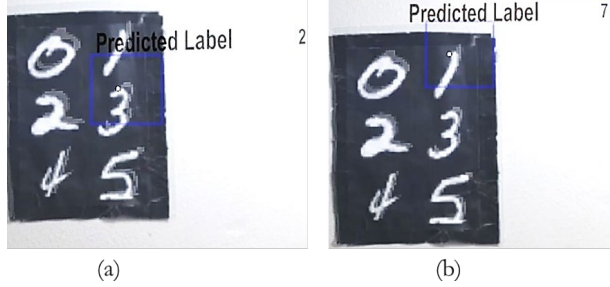

(a)        (b)

Figure 7. Examples of incorrect predictions due to unstable image acquisition: (a) Incorrect prediction where the top part of the image shows '1' and the bottom part shows '3'. (b) Incorrect prediction of '7' instead of '1' because a white horizontal line on the wall makes '1' appear similar to '7'.

### D. result and discussion

For our testing, we trained and used a LeNet-5 model, which achieved an accuracy of 98% on the MNIST test set of 1000 images. This accuracy is consistent with the expected performance of LeNet-5, indicating that the model performs reliably when applied to the digit recognition task as shown in Figure 8. The confusion matrix, depicted in Figure 8(a), provides a detailed view of the model's performance across all digit classes, further supporting our findings. Additionally, the ROC curves in Figure 8(b) illustrate the model's high performance. The ROC curves, which plot the true positive rate against the false positive rate, showed near-perfect results for each digit.

In theory, the accuracy of digit recognition using LeNet-5 should not be affected by the AR headset platform but rather by the performance of the CNN algorithm and its accuracy on the test dataset. Therefore, the integration of LeNet-5 with the HoloLens 2 does not degrade the model's predictive capability, as the headset platform's role is primarily in capturing and displaying the images, while the CNN algorithm handles the actual recognition task. However, practical factors such as image acquisition stability and environmental conditions can introduce challenges that are not present in a controlled testing environment. The experiment provided insights into both the strengths and limitations of deploying CNNs in AR environments.

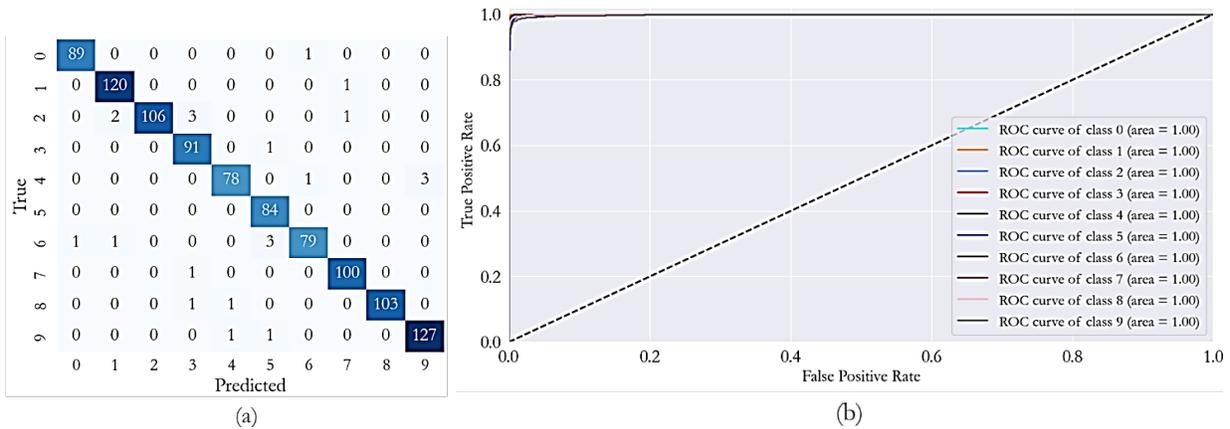

(a)        (b)

Figure 8. The result of the CNN model (a) confusion matrix (b) ROC curves

### E. Adaptation for Real-world Applications

To apply this method in real-world industrial settings, the model can be trained on various datasets with 10 labels, similar to MNIST. This training process can be executed using the `main.py` script available at https://github.com/KaMa85/LeNet. After training LeNet with the new dataset, the Python code generates a C# script that includes the optimal weights in single-dimensional arrays. These weights are then copied and replaced in the `LeNet_Kaveh.cs` file, which is available at …/LeNet/blob/main/LeNetAR/Assets/Scripts/LeNet_Kaveh.cs. Finally, the Unity project available at https://github.com/KaMa85/LeNet can be deployed on an AR headset, enabling the implementation of the application. For training the CNN model on a training set with a different number of labels and implementing it in an AR headset, the most convenient approach is to change the last layer of the LeNet model in `main.py` and `LeNet_Kaveh.cs` to adapt it to the new number of labels.

## IV. CONCLUSION

This paper presents a method to deploy CNN models on AR headsets by training them on computers and transferring optimized weight matrices. The approach transforms image data and CNN layers into a one-dimensional format suitable for the AR platform. Using the LeNet-5 model trained on the MNIST dataset with PyTorch, deployed on a HoloLens AR headset. This integration enables real-time image processing on AR headsets, enabling human input into deep AI models. While the accuracy of the LeNet-5 model remained theoretically consistent with its performance on the test set, practical issues such as image acquisition instability highlighted areas for further investigation. Ensuring model handling of unstable and low-quality image capture is crucial for maintaining recognition accuracy in real-world AR applications. Future work may focus on improving image preprocessing and stabilization techniques to mitigate the impact of environmental factors, thereby enhancing the overall reliability of the system in various practical settings.

## V. ACKNOWLEDGEMENT

We very much appreciate the help of Dr. Ji Won Chong in the experiments and also for his insights about the model. We also thank Dr. Xinxing Yuan for her valuable help during the experiments. We thank Amr Kassem and John-Wesley Hanson for their help in building the features of the Unity project.